\documentclass[letterpaper]{article} 
\usepackage{aaai2026}  
\usepackage{times}  
\usepackage{helvet}  
\usepackage{courier}  
\usepackage[hyphens]{url}  
\usepackage{graphicx} 
\usepackage{natbib}  
\usepackage{caption} 
\usepackage{algorithm}
\usepackage{algorithmic}
\usepackage{placeins}

\usepackage{newfloat}
\usepackage{listings}
\DeclareCaptionStyle{ruled}{labelfont=normalfont,labelsep=colon,strut=off} 
\floatstyle{ruled}
\newfloat{listing}{tb}{lst}{}
\floatname{listing}{Listing}
\nocopyright 

\title{Hierarchical AI-Meteorologist: LLM-Agent System\\ for Multi-Scale and Explainable Weather Forecast Reporting}
\author{
    Daniil Sukhorukov\textsuperscript{\rm 1, \rm 2},
    Andrei Zakharov\textsuperscript{\rm 1},
    Nikita Glazkov\textsuperscript{\rm 1, \rm 3},
    Katsiaryna Yanchanka\textsuperscript{\rm 4},
    Vladimir Kirilin\textsuperscript{\rm 4},
    Maxim Dubovitsky\textsuperscript{\rm 4},
    Roman Sultimov\textsuperscript{\rm 5},
    Yuri Maksimov\textsuperscript{\rm 6},
    Ilya Makarov\textsuperscript{\rm 1, \rm 7, \rm 8}
}
\affiliations{
    \textsuperscript{\rm 1}AI Research Institute, Moscow, Russia\quad
    \textsuperscript{\rm 2}ITMO University, Saint-Petersburg, Russia \quad
    \textsuperscript{\rm 3}NUST MISIS, Moscow, Russia\quad
    \textsuperscript{\rm 4}Independent Researcher \quad
    \textsuperscript{\rm 5}AI Center, Lomonosov Moscow State University, Moscow, Russia \quad
    \textsuperscript{\rm 6}LLC Interdata \quad
    \textsuperscript{\rm 7}Research Center of the Artificial Intelligence Institute, Innopolis University, Innopolis, Russia\quad
    \textsuperscript{\rm 8}ISP RAS, Moscow, Russia \quad
}

\usepackage{bibentry}

\begin{document}

\maketitle


\begin{abstract}
We present the Hierarchical AI-Meteorologist, an LLM-agent system that generates explainable weather reports using a hierarchical forecast reasoning and weather keyword generation. Unlike standard approaches that treat forecasts as flat time series, our framework performs multi-scale reasoning across hourly, 6-hour, and daily aggregations to capture both short-term dynamics and long-term trends. Its core reasoning agent converts structured meteorological inputs into coherent narratives while simultaneously extracting a few keywords effectively summarizing the dominant meteorological events. These keywords serve as semantic anchors for validating consistency, temporal coherence and factual alignment of the generated reports. Using OpenWeather and Meteostat data, we demonstrate that hierarchical context and keyword-based validation substantially improve interpretability and robustness of LLM-generated weather narratives, offering a reproducible framework for semantic evaluation of automated meteorological reporting and advancing agent-based scientific reasoning.
\end{abstract}

\section{Introduction}

Automating the interpretation of tabular, hourly weather forecasts for specific locations remains a nontrivial challenge at the intersection of meteorology and data-to-text generation. Prior works in weather Natural Language Generation (NLG), including the SumTime projects, established the importance of content selection and lexical choice when translating multivariate time series into coherent text conclusions and provided parallel "data $\leftrightarrow$ description" corpora \cite{Reiter2005,SumTimeCorpus,Belz2008}. In operational practice, National Weather Service forecasters’ Area Forecast Discussions (AFD) serve as reference texts that articulate causal reasoning and confidence levels \cite{NWS_AFD_Guide,NWS_API}. Yet a gap persists between dense numerical tables and human-readable reports, where causal links and verifiability are critical.

Despite substantial progress in machine-learning-based weather prediction, exemplified by recent systems targeting medium-range horizons \cite{GraphCast2023,PanguWeatherNature2023,WeatherBench2_2024,DLFoundationsWeather2025}, a practical question arises: how can their tabular outputs be transformed into explainable and verifiable textual reports? In this paper we treat such models purely as sources of forecast tables and intentionally avoid the modeling details. Instead, our focus is on the interpretation of these weather forecasts. Even applying LLMs and VLMs to meteorological tasks, from imagery interpretation to risk communication \cite{Lawson2025_VLM_Met,GPTCast_GMD_2025,TimeLLM_2024,EWED_VQA_2024}, are not able to  provide multi-scale interpretations of numerical data across several time resolutions at once. The need in interpretation is especially relevant for medium-range horizons beyond five days, where large tables of detailed data confound local fluctuations with dayly trends, requiring causal interpretability and readability.

In this work, we introduce the Hierarchical AI-Meteorologist, an LLM-agent pipeline that performs hierarchical interpretation of forecast tables at three concurrent levels: hourly (local dynamics), six-hourly (mesoscale patterns and noise smoothing), and daily (persistent trends and synoptic transitions). The resulting report is organized as a single narrative with consistent events and explanations across different levels. A key element of the proposed framework is the synthesis of weather keywords, a compact set of three to five terms or phrases that summarize dominant weather states and their evolution for the target time interval. For extraction and consistency control, we adapt robust keyword/keyphrase methods to the meteorological domain \cite{Mihalcea2004}. A second element is the proof-block, a brief structured “evidential” insert enumerating table-derived signals (pressure tendencies, wind shifts and strengthening, daily temperature amplitudes, precipitation duration and intensity) that support each generated keyword. This coupling supplies semantic anchors and verifiability since a stated event must manifest in observable aggregates and patterns.

The system operates on open data sources: hourly forecast tables from OpenWeather (One Call 3.0) and climatological background from Meteostat (monthly aggregates and climate normals), enabling scalable application to diverse locations without model fine-tuning \cite{OpenWeather_Guide,Meteostat_MonthlyAPI,Meteostat_NormalsAPI}. For the United States we additionally consult AFD as a weak consistency reference without requiring textual overlap \cite{NWS_AFD_Guide,NWS_API}. On the LLM side, we rely on in-context serialization of numerical data and prompts rather than task-specific training, lowering deployment barriers and improving reproducibility.

This work makes the following key contributions:
\begin{itemize}
    \item a hierarchical scheme for interpreting tabular forecasts (hourly $\to$ six-hourly $\to$ daily) and composing a causally consistent cross-scale narrative;
    \item the introduction of weather keywords as a semantic validation layer and concise summary, linked to the data through a structured proof-block;
    \item a practical, reproducible integration of open sources (OpenWeather, Meteostat) and operational texts (AFD) into an in-context LLM interpretation pipeline.
\end{itemize}

\section{Related Work}

\paragraph{LLM-agent systems for scientific reasoning.} Recent work has demonstrated the emergence of LLM-agent systems as a new paradigm for scientific reasoning. Early systems such as AutoGPT \cite{autogpt} and MetaGPT \cite{hong2023metagpt} introduced autonomous multi-agent collaboration frameworks, while AutoGen \cite{wu2024autogen} formalized conversational agent orchestration for complex analytical tasks. In scientific domains, ChemCrow \cite{m2024augmenting} and AI Scientist \cite{lu2024ai} demonstrated how LLM agents can autonomously design experiments, retrieve literature, and validate hypotheses. These systems collectively illustrate the growing shift from static language models to interactive, tool-augmented scientific agents capable of structured reasoning and knowledge generation.

\paragraph{LLMs in meteorology and weather forecasting.}
Recent work has begun leveraging LLMs to translate structured meteorological data into human-interpretable weather narratives and interactive tools. For example, ECMWF’s DestinE chatbot \cite{ecmwf_destine_chatbot_2025} to make high-resolution weather and climate data accessible via conversational interfaces, CLLMate \cite{li2024cllmate} to enable event-based forecasting of weather and climate phenomena in text form, and GPT-based forecasting \cite{franch2025gptcast} to perform precipitation nowcasting by tokenizing radar imagery and generating ensemble forecasts via language-model driven frameworks. More recent works have explored LLM-agent frameworks that integrate iterative querying, reflection, and domain-specific validation to improve the scientific accuracy and robustness of weather reports \cite{varambally2025zephyrus}.

\section{Methodology}
\begin{figure*}[h]
    \centering
    \includegraphics[width=0.98\linewidth]{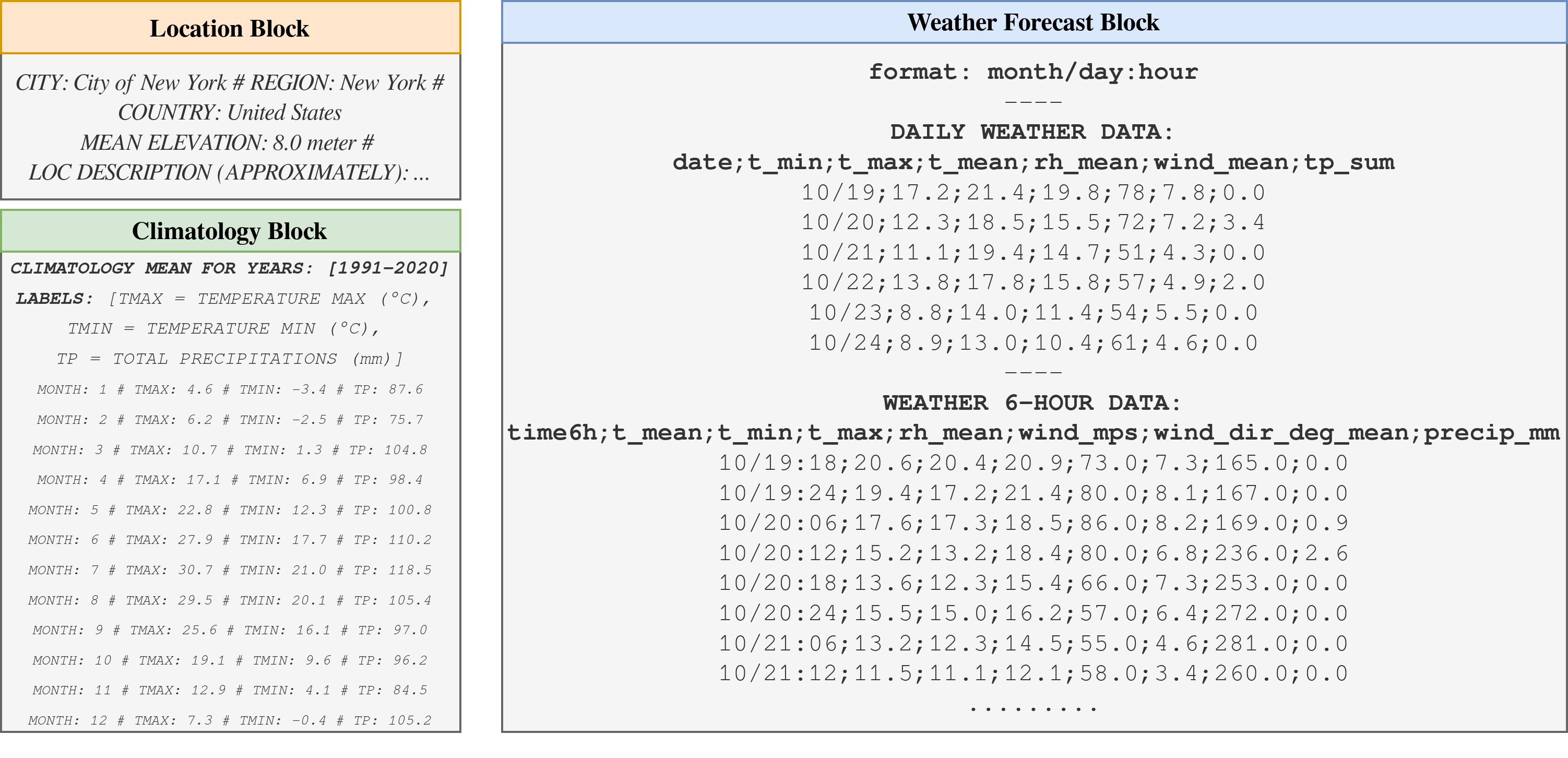}
    \caption{Schematic representation of the contextual weather data provided to the system as input for hierarchical report generation.}
    \label{fig:context_example}
\end{figure*}

\paragraph{Data acquisition.}
All inputs are assembled by the non-LLM \emph{Assistant} block. Location metadata includes city, administrative region, country, elevation above mean sea level, and an optional short description from Wikipedia when available. Climatological context (an example depicted in Fig.\ref{fig:context_example} in SM) is retrieved primarily from Meteostat (monthly aggregates and climate normals); when Meteostat is unavailable, we fall back to an ERA5-based monthly climatology on a $0.25^\circ$ grid with nearest-neighbor extraction for the requested point. For each location there are monthly $\{\mathrm{T_{min}}, \mathrm{T_{max}}, \mathrm{P_{tot}}\}$ (minimum/maximum air temperature and total precipitation). Hourly forecasts are obtained from OpenWeather One Call (time grid $\Delta t=1$\,h) and include forecast timestamp, categorical weather condition, near-surface air temperature $T$ [$^\circ$C], feeling temperature $T_{\mathrm{feel}}$ [$^\circ$C], dew point $T_d$ [$^\circ$C], relative humidity $\mathrm{RH}$ [\%], wind speed $U$ [m\,s$^{-1}$], wind direction $\theta$ [deg], wind gust $G$ [m\,s$^{-1}$], liquid/solid precipitation amount $P$ [mm], and horizontal visibility $\mathrm{Vis}$ [m]. To improve textual outputs downstream, a compact weather category is also assigned via rule-based thresholds consistent with OpenWeather condition codes.

\paragraph{Hierarchical temporal aggregation.}
To support multi-scale interpretation while controlling context length, the Assistant block forms two groups of aggregates over non-overlapping windows $W\!\in\!\{\mathrm{6h},\mathrm{1d}\}$:
\[
\overline{T}_W=\frac{1}{|W|}\sum_{t\in W}T_t,\quad
T^{\max}_W=\max_{t\in W}T_t,\quad
T^{\min}_W=\min_{t\in W}T_t,
\]
\[
\overline{\mathrm{RH}}_W=\frac{1}{|W|}\!\sum_{t\in W}\mathrm{RH}_t,\quad
\overline{U}_W=\frac{1}{|W|}\!\sum_{t\in W}U_t,\quad\\
P_W=\!\sum_{t\in W}P_t.
\]
Wind direction is averaged on the circle,
\[
\bar{\theta}_W= atan2\!\big(\frac{1}{|W|}\!\sum_{t\in W}\sin\theta_t,\;\frac{1}{|W|}\!\sum_{t\in W}\cos\theta_t\big)\;\,[^\circ],
\]
and is stored for 6-hour windows; for daily windows we omit $\bar{\theta}_{\mathrm{1d}}$ to avoid misleading circular averages. Dew point and visibility are aggregated as means; winds are summarized by mean and (optionally) maxima. Thus each 6-hour record contains \{\,$\overline{T}$, $T^{\max}$, $T^{\min}$, $\overline{\mathrm{RH}}$, $\overline{U}$, $\bar{\theta}$, $P$\,\}; each daily record mirrors this set except for wind direction.
\begin{figure}[t]
    \centering
    \includegraphics[width=0.98\linewidth]{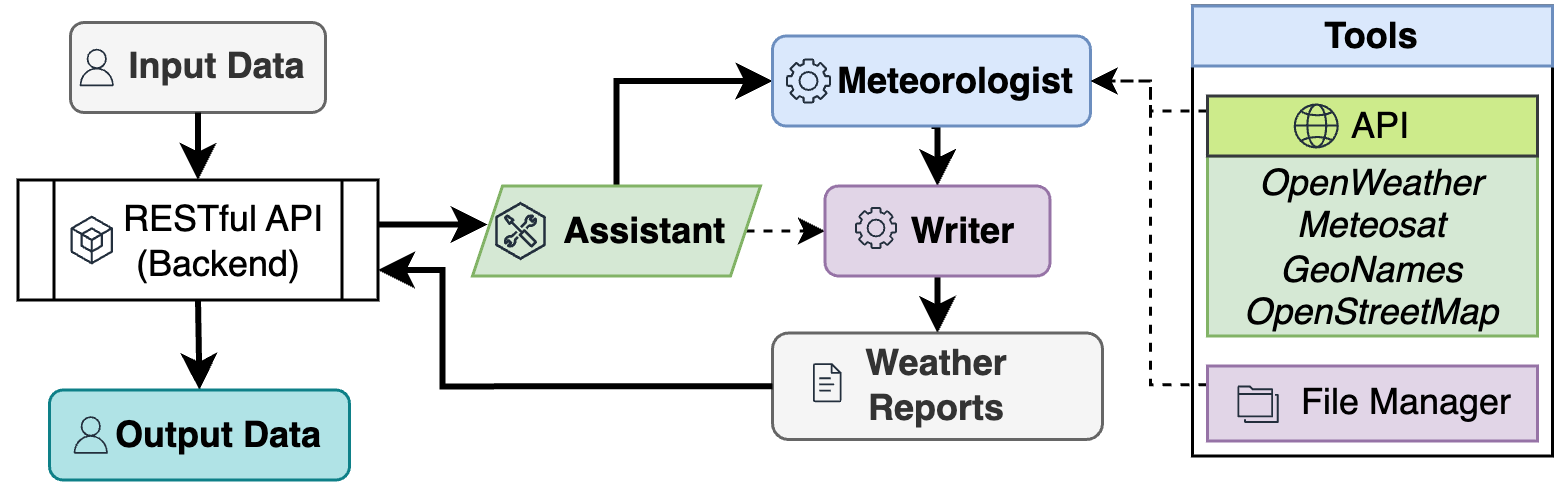}\\
    \vspace{10 pt}
    \includegraphics[width=0.98\linewidth]{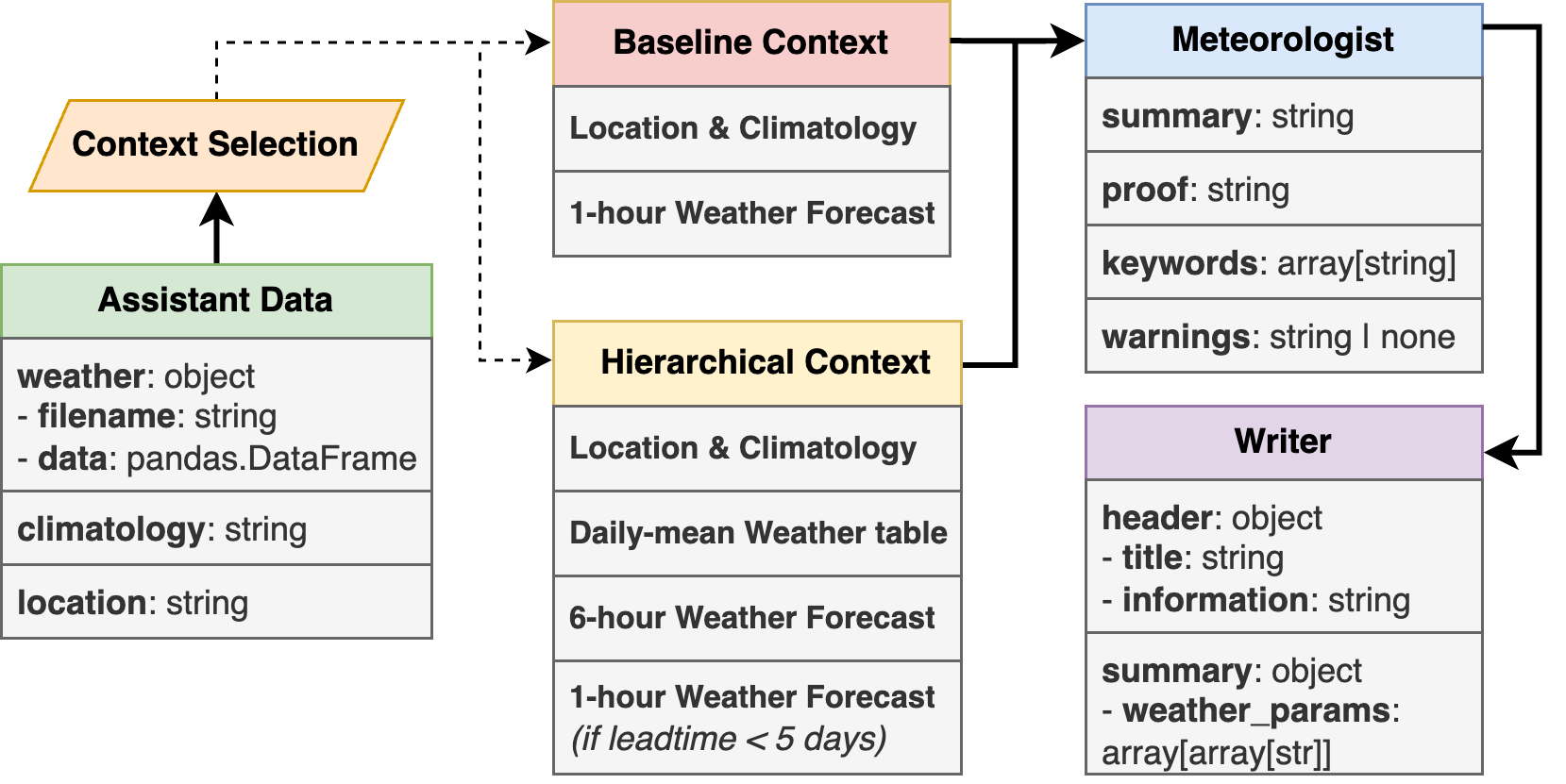}
    \caption{Overall architecture of the  Hierarchical AI-Meteorologist combining three key blocks (Assistant, Meteorologist, Writer) to automatically generate coherent weather reports. }
    \label{fig:scheme}
\end{figure}

\paragraph{Report formation and output structure.}
The \emph{Meteorologist} agent uses available context and generates a structured analysis with four fields: \texttt{summary} (multi-paragraph narrative grounded in the supplied tables), \texttt{proof} (a compact evidential rationale that points to table-derived patterns such as pressure tendencies, wind shifts/strengthening, daily temperature amplitudes, and precipitation duration/intensity), \texttt{keywords} (a list of 3-5 weather descriptors summarizing dominant states and transitions over the forecast horizon), and optional \texttt{warnings} (flagging anomalous or hazardous conditions). The downstream \emph{Writer} agent adapts this analysis to the user’s domain and style preferences and returns a JSON report over REST with a minimal, explicit format. A typical response includes
\begin{itemize}
\item \texttt{header:\{title, information\}} for location-aware titling and a brief preamble,
\item \texttt{analysis:\{summary, proof, keywords, warnings\}} ,
\item \texttt{context:\{mode, daily, six\_hour, hourly, climatology, location\}} showing the actually used tables.
\end{itemize}
This representation makes the narrative consistent because every declared keyword is expected to be supported by at least one entry in the proof block and by observable aggregates in the corresponding tables. \\


Next, we describe the proposed framework in terms of the data-processing pipeline, the construction of multi-level context, and the two-step reasoning procedure (Meteorologist $\rightarrow$ Writer). Illustrative diagram of the overall architecture and of context flows between agents are provided in the Figure \ref{fig:scheme} with the main blocks and agent links.

\subsection{System overview}
The pipeline follows a microservice (RESTful) paradigm. The external-data processing block \textit{Assistant} collects and normalizes location-specific inputs across meteorology, climatology, and geodata, then builds aggregates (6-hour and daily windows), and packages one of two context modes (baseline or hierarchical). The \textit{Meteorologist} block is an LLM agent with a structured output (\texttt{summary}, \texttt{proof}, \texttt{keywords}, \texttt{warnings}) that interprets tabular data and captures causal relations. The \textit{Writer} block is an LLM agent for post-editing to the target domain style and report format; it does not alter the factual basis produced by the Meteorologist agent, but adapts it to the user’s stylistic request, returning a JSON report together with the context, targeted to the user domain (risk analysis, energy, extreme weather).

\subsection{Input collection and normalization (Assistant)}
\paragraph{Location.} We extract city name, administrative region, country, elevation above the mean sea level. When available, a concise Wikipedia description is included in the report preamble.
\paragraph{Climatology.} The primary source is Meteostat (monthly aggregates and normals). When services are unavailable, the system returns back to monthly ERA5 climatology on a $0.25^\circ$ grid with nearest-neighbor extraction at the query point. For each month we store $\{\mathrm{T_{min}},\mathrm{T_{max}},\mathrm{P_{tot}}\}$.
\paragraph{Hourly forecast.} From OpenWeather (One Call 2.5) we use table data with $\Delta t=1$\,h consisting a timestamp, categorical “weather icon/state”, $T$ (air temperature), $T_{\mathrm{feel}}$ (feels-like), $T_d$ (dew point), $\mathrm{RH}$ (relative humidity), $U$ (wind speed), $\theta$ (wind direction), $G$ (gust), $P$ (precipitation), and $\mathrm{Vis}$ (visibility). To enhance textual outputs we additionally assign a compact weather category based on threshold rules consistent with OpenWeather codes, i.e., derived from thresholded assessments of the meteorological variables.

\subsection{Multi-level aggregation and context modes}
To enable interpretation at multiple time scales while controlling context length, the \textit{Assistant} aggregates over non-overlapping windows $W\in\{6\mathrm{h},1\mathrm{d}\}$ using means/minima/maxima for $T$, means for $\mathrm{RH}$ and $U$, sums for $P$. Wind direction is averaged on the circle and stored only for 6-hour windows. The resulting modes are:
\begin{itemize}
  \item \textbf{Baseline Context}: location and climatology plus the full hourly table (short-range forecasts, no hierarchy).
  \item \textbf{Hierarchical Context}: location and climatology, a daily table, and a 6-hour aggregate table for the location; when the lead time $H<5$ days, the hourly table is additionally included. For $H\in[5,10]$ days the hourly grid is omitted to save tokens and reduce LLM attention bias toward large tables.
\end{itemize}
Both modes are serialized into a single payload and cached (replay without repeated external API calls; ability to run different “meteorologists” over the same frozen sample).

\subsection{Stage 1: Interpreting tabular data (Meteorologist)}
\paragraph{Output schema.} The agent uses available context and returns a structured analysis consisting:
\begin{itemize}
  \item \texttt{summary} — several paragraphs describing the weather dynamics across the horizon, based on the supplied tables (daily/$6$h/$1$h);
  \item \texttt{proof} — a compact block listing observable patterns (pressure tendencies, wind shifts/strengthening, daily amplitude of $T$, duration/intensity of precipitation, etc.) that explain the observed events; the goal is to make the report verifiable and explainable for meteorological experts;
  \item \texttt{keywords} — 3–5 \emph{weather keywords} summarizing dominant weather states and transitions over the time interval (e.g., “cooling; brief rain; wind strengthening”). Generation is performed by the LLM with a controlled vocabulary and guidance to rely on aggregates; each keyword is expected to correspond to at least one feature in the \texttt{proof};
  \item \texttt{warnings} — optional anomalous/hazardous phenomena (strong winds, intense precipitation, icing, etc.) with brief data-grounded justification, when the model flags expected hazard.
\end{itemize}

\subsection{Stage 2: Domain/style adaptation (Writer)}
The \textit{Writer} receives the structured analysis and user parameters (tone, length, application domain) and composes a report with a predefined JSON structure consisting:
\begin{itemize}
  \item \texttt{header:\{title, information\}} — a location-aware title and short preamble;
  \item \texttt{analysis:\{summary, proof, keywords, warnings?\}} — the substantive content forwarded from the \textit{Meteorologist} with minimal stylistic edits;
  \item \texttt{context:\{mode, daily, six\_hour, hourly?, climatology, location\}} — an echo block listing the tables actually used to generate the text.
\end{itemize}
The Writer does not change facts from the Meteorologist. Its primary role is to adapt exposition and layout to the target user (e.g., power engineer, urban planner, agronomist).

\subsection{RESTful integration and reproducibility}
The part is split into two components: \textit{Analysis} and \textit{Report}. \textit{Analysis} accepts the serialized context and returns the structured output from the \textit{Meteorologist}. \textit{Report} composes the final report from the \textit{Writer}. Caching of “raw” and aggregated contexts (OpenWeather, Meteostat/ERA5) enables reproducing reports when models and prompt templates change, without repeated network calls. Typical reliability elements include retries for external APIs, explicit degradation codes (e.g., fallback to ERA5 climatology), validation of input JSON, and completeness checks for required fields prior to passing data to LLM agents.

\subsection{Current limitations}
The system does not perform separate “tabular reasoning” with programmatic hypothesis testing; verifiability is realized through the two-step textual rationale (\texttt{summary} + \texttt{proof}) and its linkage to \texttt{keywords}. In the present version, reports are returned as JSON, and interfaces are provided for attaching agent-based components for PDF/plot generation via preconfigured environments and Python libraries.

\section{Results}

\begin{figure*}[ht]
    \centering
    \includegraphics[width=0.98\linewidth]{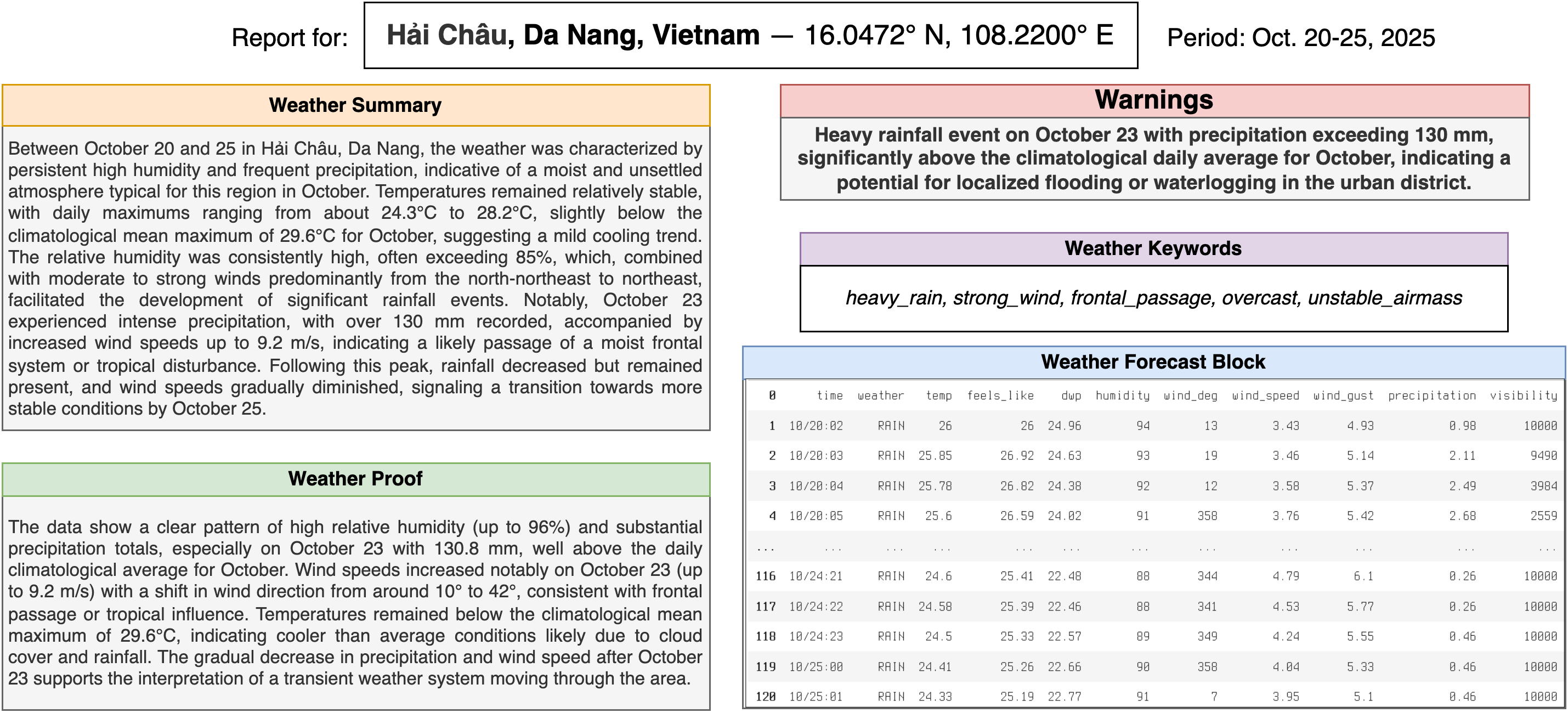}
    \caption{An example of generated report by reasoning on forecasting data with extreme events. The system automatically provides a brief summary, keywords and warnings for anticipated events. }
    \label{fig:vietnam}
\end{figure*}

We demonstrate the system performance on four different locations of distinct climate and weather behavior: Cork, Ireland ($51.903614^\circ$~N, $-8.468399^\circ$~E), Manila, Philippines ($14.5995^\circ$~N, $120.9842^\circ$~E), Chennai, India ($13.0827^\circ$~N, $80.2707^\circ$~E), and Hai Châu, Da Nang, Vietnam ($16.0472^\circ$~N, $108.2200^\circ$~E). In all cases the forecast horizon is $\sim$5–6 days (120~hours) in late October, 2025. Reports are generated in the hierarchical mode (daily + 6h; hourly rows added for short sub-intervals). We evaluate three key characteristics of report quality: (i) consistency of the \texttt{summary} with tabular aggregates, (ii) alignment of \texttt{keywords} with observed patterns, and (iii) adequacy of \texttt{proof}/\texttt{warnings} relative to the given data and climatology.

\begin{figure*}[t]
    \centering
    (a)
    \includegraphics[width=0.98\linewidth]{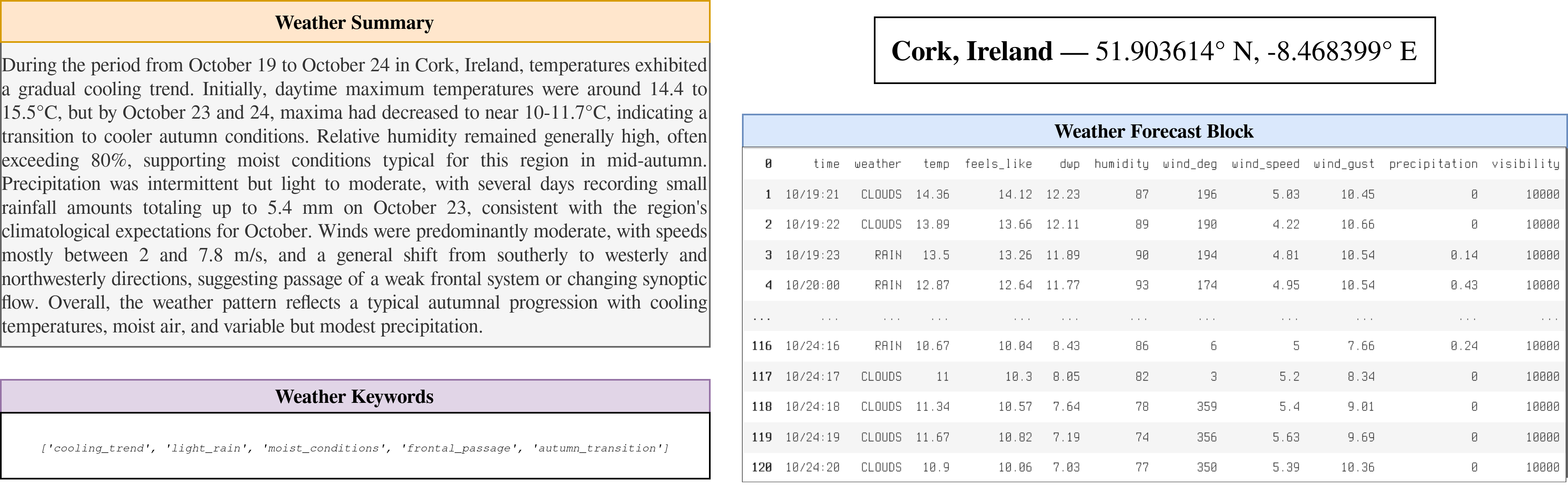}\\
    (b)\\
    \includegraphics[width=0.98\linewidth]{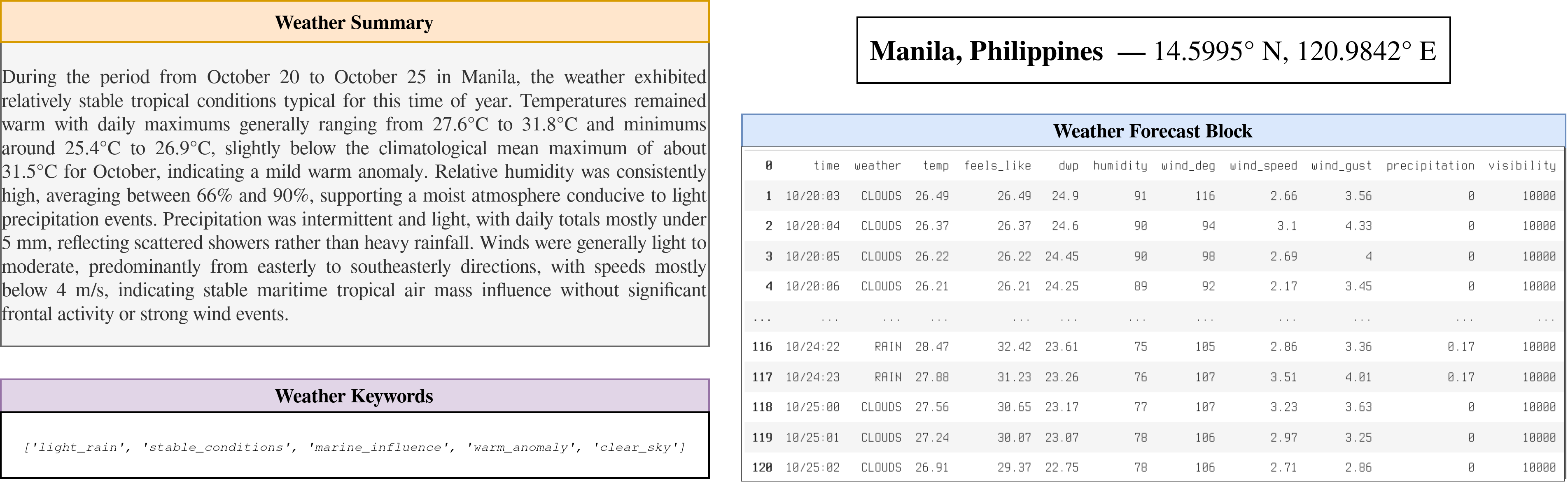}\\
    (c)\\
    \includegraphics[width=0.98\linewidth]{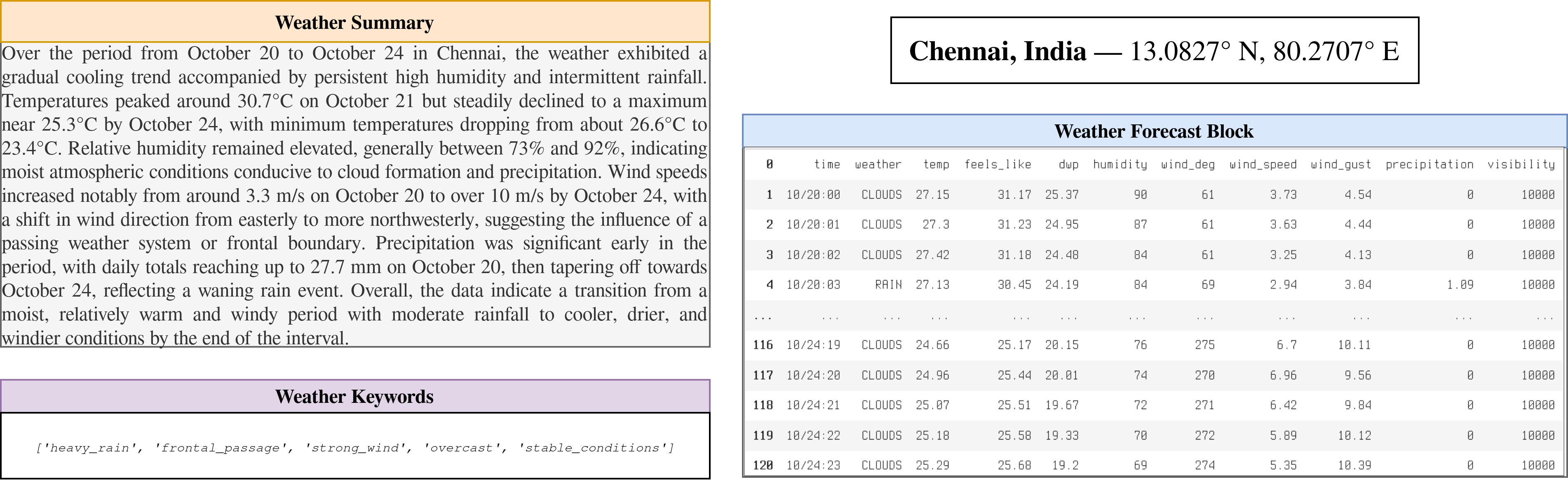}\\
    \caption{Illustrative examples of generated weather reports for three different geographic locations, demonstrating the system's ability to tailor narratives to regional weather conditions.}
    \label{fig:examples}
\end{figure*}

\paragraph{Cork, Ireland (fall transition in a mild coastal climate).}
The summary (Fig.\ref{fig:examples}a in SM) effectively captures a smooth cooling trend within forecast: daytime maxima decrease from $\approx$14.4–15.5$^\circ$C early in the window to $\approx$10–11.7$^\circ$C by October,~23–24; relative humidity often exceeds 80\%; precipitation is intermittent and light (daily total up to 5.4~mm on October,~23). Winds are moderate (2–7.8~m\,s$^{-1}$) with a shift from S to W–NW, interpreted as passage of a weak frontal disturbance.  The generated keywords \texttt{keywords} = \textit{cooling\_trend, light\_rain, moist\_conditions, frontal\_passage, autumn\_transition} compactly reflect the dominant evolution and are supported by aggregates of $T$ (declining maxima), $\mathrm{RH}$ (elevated means), $P$ (small totals), and the wind-direction shift. The narrative is readable with no false alarms; \texttt{warnings} did not trigger.

\paragraph{Manila, Philippines (persistent tropical regime with coastal influence).}
The report (Fig.\ref{fig:examples}b in SM) describes identified climatologically typical warm and humid conditions: maxima $\approx$27.6–31.8$^\circ$C, minima $\approx$25.4–26.9$^\circ$C; humidity 66–90\%; precipitation infrequent and light (daily usually $<5$~mm); winds light to moderate, predominantly E–SE, commonly $<4$~m\,s$^{-1}$. \texttt{keywords} = \textit{light\_rain, stable\_conditions, marine\_influence, warm\_anomaly, clear\_sky} are consistent with small rainfall totals, weak winds, and a coastal regime; a local ambiguity may arise for \textit{warm\_anomaly} when maxima are slightly below the climatological mean—this highlights sensitivity of the keyword to the chosen reference and thresholds. Otherwise, agreement between daily/6-hour aggregates and the \texttt{summary} is stable. \texttt{warnings} are not flagged, which perfectly corresponds to a non-extreme scenario.

\paragraph{Chennai, India (transition from humid/windy to cooler and drier).}
The data in report ((Fig.\ref{fig:examples}c in SM)) show gradual cooling and a reduction in rainfall toward the end of the window: maxima fall from $\approx$30.7$^\circ$C (October,~21) to $\approx$25.3$^\circ$C (October,~24), minima from $\approx$26.6$^\circ$C to $\approx$23.4$^\circ$C; humidity remains high (73–92\%). Wind speeds increase from $\approx$3.3 to $>10$~m\,s$^{-1}$ by October,~24 with direction shifting from E toward NW, interpreted as influence of a frontal-like process. Rainfall is substantial early (up to 27.7~mm daily on October,~20) with subsequent tapering. The system generates the following keywords \texttt{keywords} = \textit{heavy\_rain, frontal\_passage, strong\_wind, overcast, stable\_conditions} capturing a transition regime in the weather (“humid and windy” $\rightarrow$ “cooler and drier”); the \texttt{proof} block correctly relates these tags to elevated early $P$ totals, rising $U$ and a $\theta$ shift, and to decreasing daily $T$ maxima.

\paragraph{Hai Châu (Da Nang, Vietnam; intense rainfall and wind with active warnings).}
The summary in Fig.\ref{fig:vietnam} describes persistently high humidity and frequent precipitation, with an extreme daily total on October,~23  $P>130$~mm and increased wind speeds up to 9.2~m\,s$^{-1}$. A strong wind-direction shift (tens of degrees) is observed along with atmosphere temperatures “below the climatological maximum” for October, followed by a gradual decrease of rainfall and wind to October,~25. The generated \texttt{warnings} field is triggered, providing a valuable information about a risk of localized flooding due to increased above the climatological average daily total rain rate. The corresponding keywords are \texttt{keywords} = \textit{heavy\_rain, strong\_wind, frontal\_passage, overcast, unstable\_airmass}; the \texttt{proof} provides information about very high $\mathrm{RH}$ (up to 96\%), the extreme daily $P$, strengthening $U$, and a $\theta$ shift, making this case illustrative for validating hazardous events in the tropics.

\paragraph{Comparative analysis and observed model behavior.}
(1) Hierarchical presentation (daily + 6h) improves narrative coherence: for Cork and Chennai locations, the daily trend in $T$ is clearly separated from intraday variability, while in Da Nang a large daily extreme is not “lost” amid numerous hourly rows. (2) The \texttt{keywords}–\texttt{proof} coupling simplifies evaluation: for each location, keywords are supported by aggregates ($P$, $U$, $\theta$, $T$, $\mathrm{RH}$); in Manila, borderline values for \textit{warm\_anomaly} indicate that thresholds and/or reference normals may require refinement for the tropics. (3) \texttt{warnings} in those locations were triggered only in the presence of explicit extreme conditions (Da Nang), consistent with the rule design requiring large deviations from climatology and/or exceeding the predefined $P$ and $U$ thresholds. (4) The content of the \texttt{summary} remains aligned with threshold-based weather labels in the forecast tables and no false descriptions are observed, and “frontal” tags (\textit{frontal\_passage}) correlate with wind shifts and the phase structure of precipitation.

Overall, these demonstrated cases show that hierarchical context, combined with controlled \texttt{keywords} and the evidential \texttt{proof} insert, yields readable and verifiable reports for a $\sim$5–6~day horizon across diverse climate zones, from mild coastal transition patterns to typical tropical regimes in the weather with extreme rainfall events.

\section{Future Work}

The system can be further improved along three potential directions: 
\begin{enumerate}
    \item \textbf{AFD-style benchmark and auto-correction.} A NOAA AFD–inspired corpus pairing forecast tables with forecaster discussions and an error taxonomy (false/missed events, wrong trend sign, keyword–aggregate mismatch, over/under-warning) can sufficiently enhance the model performance. Additionally, a critic–corrector loop will compare \texttt{summary}/\texttt{keywords}/\texttt{proof} with daily/6h/1h aggregates and apply minimal changes (rephrase the trend description, change/remove keywords, improve the warning), exposed as lightweight \textit{Validator}/\textit{Editor} services on the current REST architecture.
    \item \textbf{Ensemble-aware interpretation.} We will augment the context with distribution tables (mean/median, p10–p90, spread, exceedance fractions) such that reports can include probability-tagged keywords (e.g., \textit{heavy\_rain [40–60\%]}), enclose trends under large spread, and add graded \texttt{warnings}. This opens ways for new experiments with the model, its probabilistic calibration and decision evaluation.

    \item \textbf{ReAct tooling for targeted validation.} An aggregate-level detector will flag uncertain and dangerous events (extreme $P$, abrupt wind shifts, keyword–data worry). A ReAct agent \cite{yao2022react} will then hypothesize, \emph{zoom} to hourly rows for the flagged window, query diagnostics (gradients, run-lengths, gust quantiles) or fallback climatology, and minimally update \texttt{proof}/\texttt{keywords}/\texttt{warnings}. Tools include \textit{aggregate-checker}, \textit{hourly-fetch}, \textit{threshold-tester}, and \textit{climatology-lookup}.
\end{enumerate}

Together, these improvements aim to make the AI-Meteorologist framework self-correcting, ensemble-aware, and able to manage with uncertain intervals on demand while preserving the current microservice/REST architecture.

\section{Conclusion}

Our experiments show that encoding forecast context at multiple temporal scales and linking concise, data-grounded keywords to an evidential proof block materially improves the usability and verifiability of LLM-generated weather narratives. Across four geographically and climatologically distinct case studies (Cork, Manila, Chennai, Da Nang) we found systm improved performance in (i) narrative–table consistency, (ii) keyword–aggregate alignment, and (iii) the relevance of proof/warning items—most notably, the system flagged and justified a hazardous rainfall/wind episode in Da Nang while avoiding false alarms in non-extreme cases. Importantly for environmental applications, the hierarchical context also reduced token-bias toward noisy hourly rows, helping the system separate daily variability from persistent, decision-relevant trends.

While the developed system already demonstrates strong promise, several clear and readily actionable enhancements can further strengthen its operational value. It can be further improved by complementing the existing text-based verification with programmatic checks, localizing keyword thresholds to improve tropical and regional sensitivity, and incorporating ensemble and uncertainty information to produce probability-aware summaries. Concrete next steps include building an AFD-style benchmark and critic–editor loop, adding ensemble-aware, probability-tagged keywords, and deploying ReAct-style tooling for targeted aggregate checks. These improvements will accelerate the system’s capabilities from a reproducible research prototype into an operational, self-checking pipeline for explainable meteorological reporting.

Collectively, this positions our Hierarchical AI-Meteorologist as a reproducible, explainable bridge between numerical forecasts and environmental decision-support. It offers interpretability, scientific rigor, and deployability for stakeholders in climate resilience, emergency management, and resource planning.

\section{Acknowledgements}
The work of I. Makarov was supported by the Ministry of Economic Development of the Russian Federation (agreement No. 139-10-2025-034 dd. 19.06.2025, IGK 000000C313925P4D0002)

\bibliography{aaai2026}

\onecolumn
\section*{Supplementary Material}

\end{document}